\relax
\documentclass[letterpaper]{article} 
\usepackage{ijcai19}  
\usepackage[colorinlistoftodos]{todonotes}
\usepackage{times}  
\usepackage{helvet}  
\usepackage{courier}  
\usepackage{url}  
\usepackage{graphicx}  
\usepackage{verbatim}
\usepackage{tabularx}
\usepackage{amssymb}
\usepackage{amsmath}
\usepackage{makecell}
\usepackage{multirow}

\begin{document}
%

\title{Controllable Neural Story Plot Generation via Reward Shaping}
\author{Pradyumna Tambwekar$^{\dagger}$\thanks{Denotes equal contribution.}, Murtaza Dhuliawala$^{\dagger*}$,  Lara J. Martin$^{\dagger}$, Animesh Mehta$^{\dagger}$,\\
{\bf\Large  Brent Harrison$^{\ddagger}$ \and Mark O. Riedl$^{\dagger}$}\\
$^{\dagger}$School of Interactive Computing, Georgia Institute of Technology\\
$^{\ddagger}$Department of Computer Science, University of Kentucky\\
}
\maketitle

\begin{abstract}
Language-modeling--based approaches to story plot generation attempt to construct a plot by sampling from a language model (LM) to predict the next character, word, or sentence to add to the story.
LM techniques lack the ability to receive guidance from the user to achieve a specific goal, resulting in stories that don't have a clear sense of progression and lack coherence.
We present a reward-shaping technique that analyzes a story corpus and produces intermediate rewards that are backpropagated into a pre-trained LM in order to guide the model towards a given goal.
Automated evaluations show our technique can create a model that generates story plots which consistently achieve a specified goal.
Human-subject studies show that the generated stories have more plausible event ordering than baseline plot generation techniques.
\end{abstract}

\section{Introduction}

{\em Automated plot generation} is the problem of creating a sequence of main plot points for a story in a given domain and with a set of specifications. 
Many prior approaches to plot generation relied on planning ~\cite{lebowitz87,gervas05,porteous2009,riedl:jair2010}. 
In many cases, these plot generators are provided with a goal, outcome state, or other guiding knowledge to ensure that the resulting story is coherent. 
However, these approaches also required extensive domain knowledge engineering.

Machine learning approaches to automated plot generation can learn storytelling and domain knowledge from a corpus of existing stories or plot summaries.
To date, most existing neural network-based story and plot generation systems lack the ability to receive guidance from the user to achieve a specific goal. 
For example, one might want a system to create a story that ends in two characters getting married.
Neural language modeling-based story generation approaches in particular~\cite{roemmele15,khalifa17,fairseq2017,Martin:aaai2018} 
are prone to generating stories with little aim since each sentence, event, word, or letter is generated by sampling from a probability distribution.
By themselves, large neural language models have been shown to work well with a variety of short-term tasks, such as understanding short children's stories~\cite{Radford2019:LM4Everything}.
However, while recurrent neural networks (RNNs) using LSTM or GRU cells can theoretically maintain long-term context in their hidden layers, in practice RNNs only use a relatively small part of the history of tokens~\cite{sharpfuzzy}.
Consequently, stories or plots generated by RNNs tend to lose coherence as the generation continues. 

One way to address both the control and the coherence issues in story and plot generation is to use reinforcement learning (RL).
By providing a reward each time a goal is achieved, a RL agent learns a policy that maximizes the future expected reward.
For plot generation, we seek a means to learn a policy model that produces output similar to plots found in the training corpus and {\em also} moves the plot along from a start state $s_0$ towards a given goal $s_g$. 
The system should be able to do this even if there is no comparable example in the training corpus where the plot starts in $s_0$ and ends in $s_g$.

Our primary contribution is a reward-shaping technique that reinforces weights in a neural language model which guide the generation of plot points towards a given goal.
{\em Reward shaping} is the automatic construction of approximate, intermediate rewards by analyzing a task domain~\cite{ng1999policy}. 
We evaluate our technique in two ways.
First, we compare our reward-shaping technique to the goal achievement rate and perplexity of a standard language modeling technique.
Second, we conduct a human subject study to compare subjective ratings of the output of our system against a conventional language modeling baseline. 
We show that our technique improves the perception of plausible event ordering and plot coherence over a baseline story generator, in addition to performing computationally better than the baseline.

\section{Background and Related Work}
Early story and plot generation systems relied on symbolic planning ~\cite{meehan77,lebowitz87,cavazza02,porteous2009,riedl:jair2010,ware11} or case-based reasoning~\cite{perez01,gervas05}. 
These techniques 
only generated stories for predetermined, well-defined domains, conflating the robustness of manually-engineered knowledge with algorithm suitability.
Regardless, symbolic planners in particular are able to provide long-term causal coherence.
Early machine learning story 
generation techniques include {\em textual} case-based reasoning trained on blogs~\cite{swanson12} and probabilistic graphical models learned from crowdsourced example stories~\cite{li:aaai2013}.

More recently, recurrent neural networks (RNNs) have been promising for story and plot generation because they can be trained on large corpora of stories and used to predict the probability of the next letter, word, or sentence in a story.
A number of research efforts have used RNNs to generate stories and plots \cite{roemmele15,khalifa17,Martin:aaai2018}.
RNNs are also often used to solve the {\em Story Cloze Test}~\cite{Mostafazadeh2016a} to predict the 5th sentence of a given story; our work differs since we focus on the generation of the entire story to fit a specified story ending. 
Similar to our task, Fan et al.~\shortcite{fairseq2018} and Yao et al.~\shortcite{yao2019planwrite} focus on a form of controllability in which a theme, topic, or title is given.

Reinforcement learning (RL) addresses some of the issues of preserving coherence for text generation when sampling from a neural language model. Additionally, it provides the ability to specify a goal. 
Reinforcement learning ~\cite{RL} is a technique that is used to solve a {\em Markov decision process} (MDP). 
An MDP is a tuple $M=\langle S, A, T , R, \gamma \rangle$ where $S$ is the set of possible world states, $A$ is the set of possible actions, $T$ is a transition function $T: S \times A \rightarrow P(S)$, $R$ is a reward function $R: S \times A \rightarrow \mathbb{R}$, and $\gamma$ is a discount factor $0 \leq \gamma \leq 1$.
The result of reinforcement learning is a {\em policy} $\pi: S\rightarrow A$, which defines which actions should be taken in each state in order to maximize the expected future reward. 
The {\em policy gradient} learning approach to reinforcement learning 
directly optimizes the parameters of a policy model, which is represented as a neural network.
One model-free policy gradient approach, REINFORCE~\cite{REINFORCE}, learns a policy by sampling from the current policy and backpropagating any reward received through the weights of the policy model.

Deep reinforcement learning (DRL) has been used successfully for dialog--a similar domain to plot generation. 
Li et al.~\shortcite{li-drl} pretrained a neural language model and then used a policy gradient technique to reinforce weights that resulted in higher immediate reward for dialogue. 
They also defined a coherence reward function for their RL agent. 
Contrasted with task-based dialog generation, story and plot generation often require long-term coherence to be maintained. 
Thus we use reward shaping to force the policy gradient search to seek out longer-horizon rewards.


\section{Reinforcement Learning for Plot Generation}

We model story generation as a planning problem: find a sequence of events that transitions the state of the world into one in which the desired goal holds.
In the case of this work, the goal is that a given verb (e.g., marry, punish, rescue) occurs in the final event of the story.
While simplistic, it highlights the challenge of control in story generation.

Specifically, we use reinforcement learning to plan out the events of a story and use policy gradients to learn a policy model.
We start by training a language model on a corpus of story plots.
A language model $P(x_n|x_{n-1}...x_{n-k}; \theta)$ gives a distribution over the possible tokens $x_n$ that are likely to come next given a history of tokens $x_{n-1}...x_{n-k}$ and the parameters of a model $\theta$.
This language model is a first approximation of the policy model.
Generating a plot by iteratively sampling from this language model, however, provides no guarantee that the plot will arrive at a desired goal except by coincidence.
We use REINFORCE~\cite{REINFORCE} to specialize the language model to keep the local coherence of tokens initially learned and also to prefer selecting tokens that move the plot toward a given goal.

If reward is only provided when a generated plot achieves the given goal, then the rewards will be very sparse. 
Policy gradient learning requires a dense reward signal to provide feedback after every step. 
As such, our primary contribution is a reward-shaping technique where the original training corpus is automatically analyzed to construct a dense reward function that guides the plot generator toward the given goal.

\begin{figure}[tb]
\centering
\includegraphics[width=0.4\textwidth]{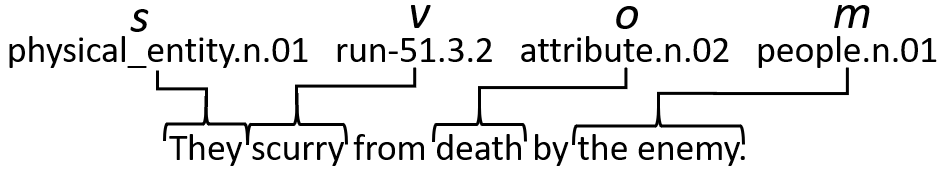}
\vspace{-0.5\baselineskip}
 \label{fig:translation}
 \caption{An example sentence and its event representation. }
 \end{figure}

\subsection{Initial Language Model}

Martin et al.~\shortcite{Martin:aaai2018} demonstrated that the predictive accuracy of a plot generator could be improved by switching from natural language sentences to an abstraction called an {\em event}. 
An event is a tuple $e=\langle wn(s), vn(v), wn(o), wn(m)\rangle$, where $v$ is the verb of the sentence, $s$ is the subject of the verb, $o$ is the object of the verb, and $m$ is a propositional object, indirect object, causal complement, or any other significant noun.
The parameters $o$ and $m$ may take the special value of {\em empty} to denote there is no object of the verb or any additional sentence information, respectively.
We use the same event representation in this work.
As in Martin et al., we stem all words and then apply the following functions.
The function $wn(\cdot)$ gives the WordNet~\cite{wordnet} {\em Synset} of the argument two levels up in the hypernym tree (i.e. the grandparent {\em Synset}).
The function $vn(\cdot)$ gives the VerbNet~\cite{verbnet} class of the argument.
See Figure~\ref{fig:translation} for an example of an ``eventified'' sentence.
Where possible, we split sentences into multiple events, 
which creates 
a potential one-to-many relationship between a 
sentence and the event(s) it produces.
Once produced, events are used sequentially. 

In this paper, we use an encoder-decoder network \cite{sutskever2014sequence} as our starting language model and our baseline. 
Encoder-decoder networks can be trained to generate sequences of text for dialogue or story generation by pairing one or more sentences in a corpus with the successive sentence and learning a set of weights that captures the relationship between the sentences.
Our language model is thus $P(e_{i+1}|e_{i};\theta)$ where $e_i=\langle s_{i}, v_{i}, o_{i}, m_{i}\rangle$.

\subsection{Policy Gradient Descent}

We seek a 
policy model $\theta$ such that 
$P(e_{i+1}|e_{i};\theta)$ 
is the distribution over events according to the corpus {\em and} also that increase the likelihood of reaching a given goal event in the future. 
For each input event 
$e_{i}$ 
in the corpus, an action involves choosing the most probable next event 
$e_{i+1}$ 
from the probability distribution of the language model. 
The reward is calculated by determining how far the event 
$e_{i+1}$
is from our given goal event. 
The final gradient used for updating the parameters of the network and shifting the distribution of the language model is calculated as follows: 
\begin{equation}
\label{eq:gradient}
\nabla_{\theta}{J(\theta)} = R(v(e_{i+1}))\nabla_{\theta}{log P(e_{i+1}|e_{i};\theta)}
\end{equation}
where 
$e_{i+1}$
and $R(v(e_{i+1}))$ are, respectively, the event chosen at timestep $i+1$ and the reward for the verb in that event.
The policy gradient technique thus gives an advantage to highly-rewarding events by facilitating a larger step towards the likelihood of predicting these events in the future, over events which have a lower reward. 
In the next section we describe how $R(v(e_{i+1}))$ is computed.

\subsection{Reward Shaping}

For the purpose of {\em controllability} in plot generation, we wish to reward the network whenever it generates an event that makes it more likely to achieve the given goal. 
For the purposes of this paper, the goal is a given VerbNet class that we wish to see at the end of a plot.
{\em Reward shaping}~\cite{ng1999policy} is a technique whereby sparse rewards---such as rewarding the agent only when a given goal is reached---are replaced with a dense reward signal that provides rewards at intermediate states in the exploration leading to the goal.

To produce a smooth, dense reward function, we make the observation that certain events---and thus certain verbs---are more likely to appear closer to the goal than others in story plots. 
For example, suppose our goal is to generate a plot in which one character {\em admires} another ({\em admire-31.2} is the VerbNet class that encapsulates the concept of falling in love).
Events that contain the verb {\em meet} are more likely to appear nearby events that contain {\em admire}, whereas events that contain the verb {\em leave} are likely to appear farther away.

%

To construct the reward function, we pre-process the stories in our training corpus and calculate two key components: (a)~the distance of each verb from the target/goal verb, and (b)~the frequency of the verbs found in existing stories.

\subsubsection{Distance}

The distance component of the reward function measures how close the verb $v$ of an event is to the target/goal verb $g$, which is used to reward the model when it produces events with verbs that are closer to the target verb. 
The formula for estimating this metric for a verb $v$ is:
\begin{equation}
r_1(v)=\log\sum_{s\in S_{v,g}}{l_s-d_s(v,g)}
\end{equation}
where $S_{v,g}$ is the subset of stories in the corpus that contain $v$ prior to the goal verb $g$, $l_s$ is the length of story $s$, and $d_s(v,g)$ is the number of events between the event containing $v$ and the event containing $g$ in story $s$ (i.e., the distance within a story). 
Subtracting from the length of the story produces a larger reward when events with $v$ and $g$ are closer.

\subsubsection{Story-Verb Frequency}

Story-verb frequency rewards based on how likely any verb is to occur in a story before the target verb. 
This component estimates how often a particular verb $v$ appears before the target verb $g$ throughout the stories in the corpus.
This discourages the model from outputting events with verbs that rarely occur in stories before the target verb.
The following equation is used for calculating the story-verb frequency metric: 
\begin{equation}
r_2(v)=\log\frac{k_{v,g}}{N_{v}}
\end{equation}
where $N_v$ is the count of verb $v$ in the corpus, and $k_{v,g}$ is the number of times $v$ appears before goal verb $g$ in any story.

\subsubsection{Final Reward}

The final reward for a verb---and thus event as a whole---is calculated as the product of the distance and frequency metrics.
The rewards are normalized across all the verbs in the corpus.
The final reward is:
\begin{equation}
\label{eq:final_reward}
R(v) = \alpha\times r_1(v)\times r_2(v)
\end{equation}
where $\alpha$ is the normalization constant. 
When combined, both $r$ metrics advantage verbs that 1)~appear close to the target, while also 2)~being present before the target in a story frequently enough to be considered significant.

\subsubsection{Verb Clustering}
\label{sec:clustering}

In order to discourage the model from jumping to the target quickly, we cluster  
all verbs based on Equation~\ref{eq:final_reward} using the Jenks Natural Breaks optimization technique~\cite{jenks1971error}. 
We restrict the vocabulary of $v_\textrm{out}$---the model's output verb---to the set of verbs in the $c+1^{th}$ cluster, where $c$ is the index of the cluster that verb $v_\textrm{in}$---the model's input verb---belongs to. 
The rest of the event is generated by sampling from the full distribution.
The intuition is that by restricting the vocabulary of the output verb, the gradient update in Equation~\ref{eq:gradient} takes greater steps toward verbs that are more likely to occur next (i.e., in the next cluster) in a story headed toward a given goal.
If the sampled verb has a low probability, the step will be smaller than 
if the verb is highly probable according to the language model.

\section{Automated Experiments}

We ran experiments to measure three properties of our model:
(1)~how often our model can produce a plot---a sequence of events---that contains a desired target verb; (2)~the perplexity of our model; and (3)~the average length of the stories.
Perplexity is a measure of the predictive ability of a model;   
particularly, how ``surprised'' the model is by occurrences in a corpus.
We compare our results to those of a baseline event2event story generation model from Martin et al.~\shortcite{Martin:aaai2018}.

\subsection{Corpus Preparation}

We use the CMU movie summary corpus~\cite{wikiplots}. 
However, this corpus proves to be too diverse; 
there is high variance between stories, which dilutes event patterns.
We used  Latent Dirichlet Analysis to cluster the stories from the corpus into 100 ``genres''.
We selected a cluster that appeared to contain soap-opera--like plots. 
The stories were ``eventified''---turned into {\em event} sequences, as explained in the {\em Initial Language Model} section of the paper.  
We chose {\em admire-31.2} and {\em marry-36.2} as two target verbs because those VerbNet classes capture the sentiments of ``falling in love'' and ``getting married'', which are appropriate for our sub-corpus.
The romance corpus was  split into 90\% training, and 10\% testing data.
We used consecutive events from the eventified corpus as source and target data, respectively, for training the sequence-to-sequence network. 

\subsection{Model Training}

For our experiments we trained the encoder-decoder network using Tensorflow.
Both the encoder and the decoder comprised of LSTM units, with a hidden layer size of 1024. 
The network was pre-trained for a total of 200 epochs using mini-batch gradient descent and batch size of 64. 

We created three models:
\begin{itemize}
\item {\em Seq2Seq}: This pre-trained model is identical to the ``generalized multiple sequential event2event'' model in Martin et al.~\shortcite{Martin:aaai2018}. This is our baseline.
\item {\em DRL-clustered}: Starting with the weights from the {\em Seq2Seq}, we continued training using the policy gradient technique and the reward function, along with the clustering and vocabulary restriction in the verb position described in the previous section, while keeping all network parameters constant. 
\item {\em DRL-unrestricted}: This is the same as {\em DRL-clustered} but without vocabulary restriction while sampling the verb for the next event during training (\S~{\em Verb Clustering}).
\end{itemize}

\noindent
The DRL-clustered and DRL-unrestricted models are trained for a further 200 epochs than the baseline.

\subsection{Experimental Setup}

With each event in our held-out dataset as a seed event, we generated stories with our baseline Seq2Seq, DRL-clustered, and DRL-unrestricted models. 
For all models, the story generation process was terminated when: 
(1) the model outputs an event with the target verb;
(2) the model outputs an end-of-story token; or 
(3) the length of the story reaches 15 lines. 

Goal achievement rate was calculated by measuring the percentage of these stories that ended in the target verb ({\em admire} or {\em marry}).
Additionally, we average generated story lengths to compare to the average story length in our test data where the goal event occurs (setting length to 15 if it doesn't occur).
Finally, we measure the perplexity for all the models, with the exception of the testing data since it is not a model. 

\subsection{Results and Discussion}

Results are summarized in Table ~\ref{tab:experiments}.
Only $22.47\%$ of the stories in the testing set, on average, end in our desired goals, illustrating how rare the chosen goals were in the corpus.
The DRL-clustered model generated the given goals on average $93.82\%$ of the time, compared to $37.72\%$ on average for the baseline Seq2Seq and $19.935\%$ for the DRL-unrestricted model. 
This shows that
our policy gradient approach can direct the plot to a pre-specified ending and that our clustering method is integral to doing so.
Removing verb clustering from our reward calculation to create the DRL-unrestricted model harms goal achievement; the system rarely sees a verb in the next cluster so the reward is frequently low, making distribution shaping towards the goal difficult. 

We use perplexity as a metric to estimate how accurate the learned distribution is for predicting unseen data. 
We observe that perplexity values drop substantially for the DRL models ($7.61$ for DRL-clustered and $5.73$ for DRL-unrestricted with goal {\em admire}; $7.05$ for DRL-clustered and $9.78$ for DRL-unrestricted with goal {\em marry}) when compared with the Seq2Seq baseline ($48.06$). 
This can be attributed to the fact that our reward function is based on the distribution of verbs in the story corpus, refining the model's ability to recreate the corpus distribution.
Because DRL-unrestricted's rewards are based on subsequent verbs in the corpus instead of verb clusters, it sometimes results in a lower perplexity, but at the expense of not learning how to achieve the goal often.

The average story length is an important metric because it is trivial to train a language model that reaches the goal event in a single step.
DRL models don't have to produce stories the same length as the those in the testing corpus, as long as the length is not extremely short (leaping to conclusions) or too long (the story generator is timing out).
The baseline Seq2Seq model creates stories that are about the same length as the testing corpus stories, showing that the model is mostly mimicking the behavior of the corpus it was trained on. 
The DRL-unrestricted model produces similar behavior, due to the absence of clustering or vocabulary restriction to prevent the story from rambling. 
However, the DRL-clustered model creates slightly shorter stories, showing that it is reaching the goal quicker, while not jumping immediately to the goal.

\setlength{\tabcolsep}{4pt}

\begin{table}[t]

\footnotesize
\centering
\begin{tabularx}{\columnwidth}{|c|X|X|X|X|}
\hline
  \rotatebox[origin=c]{90}{\bf Goal} &\thead{\bf Model} & \thead{\bf Goal\\ \bf achievement \\\bf rate} & \thead{\bf Average\\\bf perplexity} & \thead{\bf Average\\\bf story\\\bf length}\\
  \hline
  \makecell{\multirow{4}{*}{\rotatebox[origin=c]{90}{\em admire}}}&\makecell{Test Corpus} & \makecell{20.30\%} & \makecell{n/a} & \makecell{7.59}\\
  &\makecell{Seq2Seq} & \makecell{35.52\%} & \makecell{48.06} & \makecell{7.11}\\
   &\makecell{Unrestricted}  & \makecell{15.82\%} & \makecell{\bf 5.73} & \makecell{7.32}\\
& \makecell{Clustered}  & \makecell{\bf 94.29\%} & \makecell{7.61} & \makecell{4.90}\\
  \hline
  \makecell{\multirow{4}{*}{\rotatebox[origin=c]{90}{\em marry}}}&\makecell{Test Corpus} & \makecell{24.64\%} & \makecell{n/a} & \makecell{7.37}\\
  &\makecell{Seq2Seq} & \makecell{39.92\%} & \makecell{48.06} & \makecell{6.94}\\
    &\makecell{Unrestricted}  & \makecell{24.05\%} & \makecell{9.78} & \makecell{7.38}\\

  & \makecell{Clustered} & \makecell{\bf 93.35\%} & \makecell{\bf 7.05} & \makecell{5.76}\\
  \hline
  \end{tabularx}
    \caption{Results of the automated experiments, comparing the goal achievement rate, average perplexity, and average story length for the testing corpus, baseline Seq2Seq model, and our clustered and unrestricted DRL models.\label{tab:experiments}}
\end{table}

\section{Human Evaluation}
The best practice in the evaluation of story/plot generation is human subject evaluation.
However, the use of the event 
representation makes human subject evaluation difficult since events are not easily readable. 
Martin et al.~\shortcite{Martin:aaai2018} used a second neural network to translate events into human-readable sentences, but their technique did not have sufficient accuracy to use in a human evaluation.
The use of a second network also makes it impossible to isolate the generation of events from the generation of the final natural language sentence 
in terms of human perception.
To overcome this challenge, we have developed an evaluation protocol that allows us to directly evaluate plots with human judges. 
Specifically, we recruited and taught individuals to convert event sequences into natural language before giving generated plots to human judges. 
By having concise, grammatically- and semantically-guaranteed human translations of generated plot events we know that the human judges are evaluating the raw events and not the creative aspects of the way sentences are written. 

\subsection{Corpus Creation}
We collected 5 stories generated by our DRL-clustered system, 5 generated from our Seq2Seq baseline, and 3 from the eventified testing corpus.
The stories were selected by randomly picking start events---keeping the same start events across conditions---until we had stories that were 5-10 events long.
By keeping a story length limit, we guarantee having DRL stories that reached the goal.
The testing corpus was mainly used to verify the translation process's accuracy since we do not expect our models to reach this upper bound; thus only three stories were selected.
We trained 26 unbiased people to ``translate'' events into short natural language sentences.

Each translator was instructed that their ``primary goal is to produce faithful translations of stories from an abstract `event' representation into a natural language sentence.'' The instructions then continued with:
(1)
a refresher on parts of speech,
(2)
the format of the event representation,
(3)
examples of events and their corresponding sentences,
(4)
resources on WordNet and VerbNet with details on how to use both, and
(5)
additional general guidelines and unusual cases they might encounter (e.g., how to handle {\em empty} parameters in events).
The translators were further instructed 
to not add extraneous details, swap the order of words in the event, nor choose a better verb even if the plot would be improved.

Pairs of people translated plots individually and then came together to reach a consensus on a final version of the plot.
That is, human translators reversed the eventification process 
to create a human-readable sentence from an event. 
Table~\ref{tab:Story_Translation} shows an example of an entire eventified story and the corresponding human translations.

\begin{table*}[t]
\footnotesize
\centering
\begin{tabular}{|l|l|}
\hline
  {\bf Event Output {\em(subject, verb, object, modifier)}} & {\bf Translated Sentence}\\
  \hline
  relative.n.01, disappearance-48.2, $empty$, $empty$ & My cousin died.\\
  \hline
  PERSON1, say-37.7-1, visit, $empty$ & Alexander insisted on a visit.\\
  \hline
  \makecell[l]{PERSON1, meet-36.3-1, female.n.02, $empty$} & Alexander met her.\\
  \hline
  PERSON0, correspond-36.1, $empty$, PERSON1 & Barbara commiserated with Alexander.\\
  \hline
  physical\_entity.n.01, marry-36.2, $empty$, $empty$ & They hugged.\\
  \hline
  group.n.01, contribute-13.2-2, $empty$, LOCATION & The gathering dispersed to Hawaii.\\
  \hline
  gathering.n.01, characterize-29.2-1-1, time\_interval.n.01, $empty$ & The community remembered their trip.\\
  \hline
  physical\_entity.n.01, cheat-10.6, pack, $empty$ & They robbed the pack.\\
  \hline
  physical\_entity.n.01, admire-31.2, social\_gathering.n.01, $empty$ & They adored the party.\\
  \hline
  \end{tabular}
  \caption{An example eventified story from the DRL-clustered system paired with the translation written by a pair of participants.\label{tab:Story_Translation}}
\end{table*}

\subsection{Experimental Setup}

We recruited 175 participants on Amazon Mechanical Turk. 
Each participant was compensated \$10 for completing the questionnaire. 
Participants were given one of the translated plots at a time, rating each of the following statements on a 5-point Likert scale for how much they agreed (Strongly Agree, Somewhat Agree, Neither Agree nor Disagree, Somewhat Disagree, or Strongly Disagree):

\begin{enumerate}
    \item This story exhibits CORRECT GRAMMAR.
    \item This story's events occur in a PLAUSIBLE ORDER.
    \item This story's sentences MAKE SENSE given sentences before and after them.
    \item This story AVOIDS REPETITION.
    \item This story uses INTERESTING LANGUAGE.
    \item This story is of HIGH QUALITY.
    \item This story is ENJOYABLE.
    \item This story REMINDS ME OF A SOAP OPERA.
    \item This story FOLLOWS A SINGLE PLOT.
\end{enumerate}

\noindent Through 
the equal interval assumption, we turn Likert values into numerals 1 (Strongly Disagree) to 5 (Strongly Agree).

The first seven questions are taken from a tool designed specifically for the evaluation of computer-generated stories which has been validated against human judgments~\cite{purdy2018}.
Each participant answered the questions for all three story conditions.
The question about the story being a soap opera was added to determine how the performance of the DRL story generator affects reader perceptions of the theme, since the system was trained on soap-opera--like plots.
The single plot question was added to determine if our DRL model was maintaining the plot better than the Seq2Seq model.
The questions about correct grammar, interesting language, and avoiding repetition are irrelevant to our evaluation since the natural language was produced by the human translators but were kept for consistency with Purdy et al.~\shortcite{purdy2018}.

Finally, participants answered two additional prompts that required short answer responses: (1) Please give a summary of the story above in your own words; and (2) For THIS STORY, please select which of the previous attributes (e.g. enjoyable, plausible, coherent) you found to be the MOST IMPORTANT and explain WHY.
The answers to these questions were not evaluated, but if any participants failed to answer the short answer questions, their data was removed from the results.
We removed 25 participants' data in total.

\begin{figure*}[t]
\centering
	\includegraphics[width=0.75\textwidth]{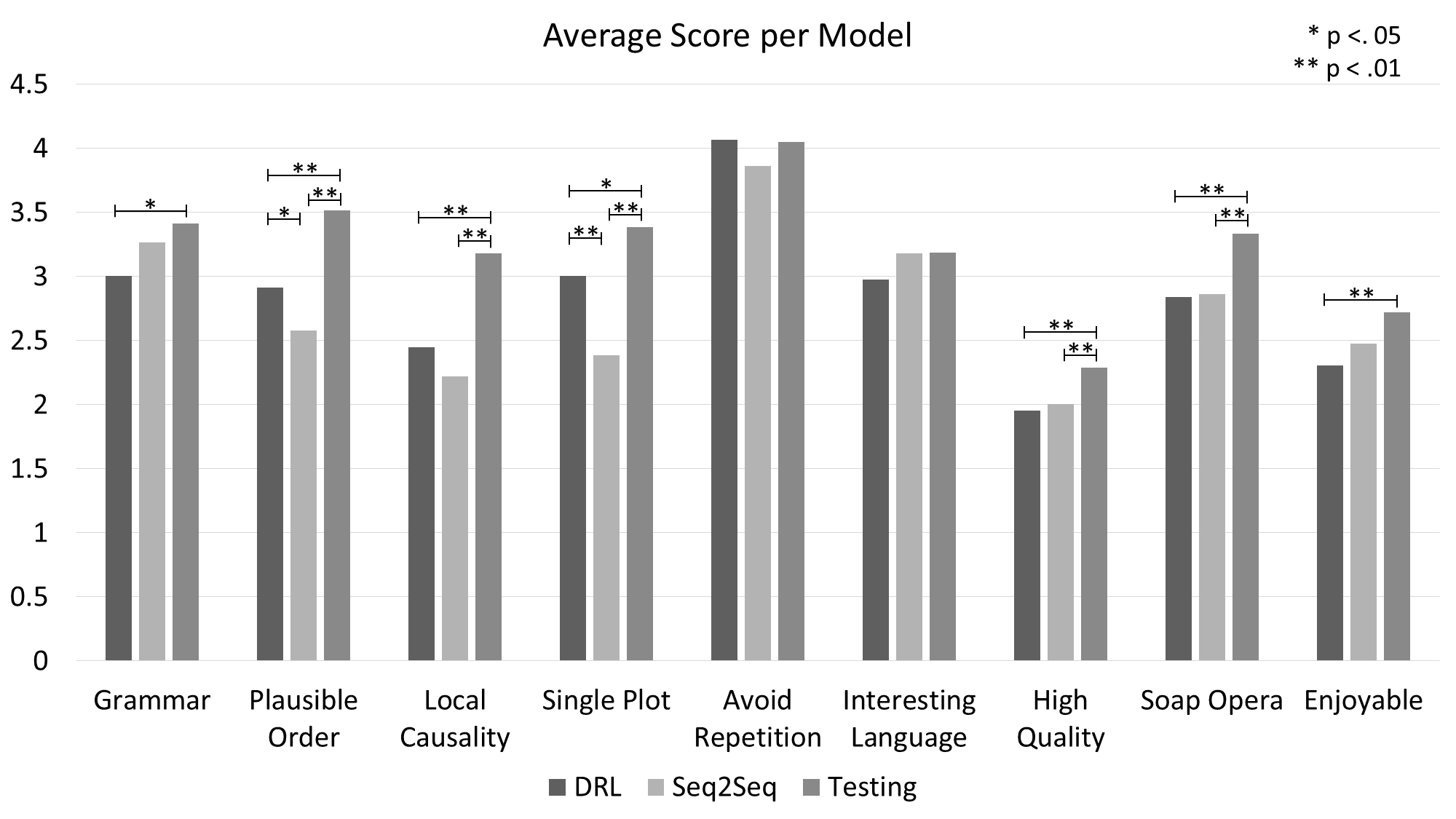}
	\caption{Participants rated a story from each category across nine different dimensions on a scale of 1-Strongly Disagree to 5-Strongly Agree. Single asterisks stand for p\textless 0.05. Double asterisks stand for p\textless 0.01.}
    \label{fig:scores}
\end{figure*}

\subsection{Results and Discussion}

We performed one-way repeated-measures ANOVA on the data since each participant rated a story from each category, using Tukey HSD as the post-test.
We verified that the data is normal, and the variances are not significantly different. The data was collected independently.
Average scores and their significance across conditions can be seen in Figure \ref{fig:scores}. 

%
Questions on interesting language and avoiding repetition are not found to be significant across all three conditions.
Since these are not related to event generation model performance this provides an indication that the translations are fair across all conditions.
Grammar was significantly different between testing corpus stories and DRL-generated stories ($p<0.05$), which was unanticipated.
Upon further analysis, both the baseline Seq2Seq model and the DRL model generated {\em empty} values for the object and modifier at higher rates than found in the corpus.
It is harder to make complete, grammatical sentences with only two tokens in an event, namely when a verb is transitive--requiring at least one object.
Beyond more expected results, such as having a better plausible order, the testing corpus stories were also significantly more likely to be perceived as being soap operas ($p<0.01$), the genre from which the corpus stories were drawn. 
It is unclear why this would be the case, except that both the Seq2Seq and DRL models could be failing to learn some aspect of the genre despite being trained on the same corpus.
It is also worth noting that randomly-selecting 5 generated stories does not guarantee that they will be representative of their respective models.

Stories in the DRL condition were significantly perceived to have more plausible orderings than those in the baseline Seq2Seq condition ($p<0.05$) and were significantly more likely to be perceived as following a single plot ($p<0.05$). 
Since stories generated by the baseline Seq2Seq model begin to lose coherence as the story progresses, these results confirm our hypothesis that the DRL's use of reward shaping keeps the plot on track.
The DRL is also perceived as generating stories with more local causality than the Seq2Seq, although the results were not statistically significant.

For all other dimensions, the DRL stories are not found to be significantly different than baseline stories.
When further training a 
pre-trained language model using a reward function instead of the standard cross-entropy loss there is a non-trivial chance that model updates will degrade any aspect of the model that is not related to goal achievement.
Thus, a positive result is one in which DRL-condition stories are never significantly lower than Seq2Seq-condition stories.
This shows that we are able to get to the goal state without any significant degradation in other aspects of story generation. 



\section{Conclusions}

Language model--based story and plot generation systems produce stories that lack direction.
%
Our reward shaping technique learns a policy that generates stories that are probabilistically comparable with the training corpus while also reaching a pre-specified goal $\sim$$93\%$ of the time. 
Furthermore, the reward-shaping technique improves perplexity when generated plots are compared to the testing corpus. 
However, in plot generation, the comparison to an existing corpus is not the most significant metric because novel plots may also be good.
A human subject study showed that the reward shaping technique significantly improves the plausible ordering of events and the likelihood of producing a sequence of events that is perceived to be a single, coherent plot.
We thus demonstrated for the first time that control over neural plot generation can be achieved in the form of providing a goal that indicates how a plot should end.



\section{Acknowledgements}
This work is supported by DARPA W911NF-15-C-0246.
The views, opinions, and/or conclusions contained in this paper are those of the authors and should not be interpreted as representing the official views or policies, either expressed or implied of the DARPA or the DoD.

\bibliographystyle{named}
\bibliography{refs}

\end{document}